\def\year{2021}\relax
\documentclass[letterpaper]{article} 
\usepackage{aaai21}  
\usepackage{times}  
\usepackage{helvet} 
\usepackage{courier}  
\usepackage[hyphens]{url}  
\usepackage{graphicx} 
\usepackage{natbib}  
\usepackage{multirow}
\usepackage{amsmath}
\usepackage{caption} 
\usepackage{diagbox}

\usepackage{microtype}
\usepackage{url}
\usepackage[font=small]{caption}
\usepackage{graphicx}
\usepackage{amsmath}
\usepackage{amsthm}
\usepackage{booktabs}
\usepackage[ruled]{algorithm2e}
\usepackage{subfiles}
\usepackage{multirow}
\usepackage{graphicx}
\usepackage{amsfonts}
\usepackage{bbm}
\usepackage{url}
\usepackage[noend]{algpseudocode}
\usepackage{arydshln}
\usepackage{booktabs}
\usepackage{color}
\usepackage{subfigure}
\usepackage{colortbl}
\usepackage{enumitem}
\usepackage[symbol]{footmisc}

\usepackage{xcolor}
\newcommand{\nop}[1]{}

\title{Reinforced Multi-Teacher Selection for Knowledge Distillation}

\author {
{Fei Yuan}\textsuperscript{\rm 1}\thanks{Work is done during internship at Microsoft STCA NLP Group.} 
{ Linjun Shou}\textsuperscript{\rm 2}
{ Jian Pei}\textsuperscript{\rm 3} 
{ Wutao Lin}\textsuperscript{\rm 2}
{ Ming Gong}\textsuperscript{\rm 2} 
{ Yan Fu}\textsuperscript{\rm 1}
{ Daxin Jiang}\textsuperscript{\rm 2}\thanks{Corresponding author.}\\
}
\affiliations {
  \textsuperscript{\rm 1} {University of Electronic Science and Technology of China} \\
  \textsuperscript{\rm 2} {Microsoft STCA NLP Group} \\
  \textsuperscript{\rm 3} {School of Computing Science, Simon Fraser University} \\
  \texttt{feiyuan@std.uestc.edu.cn}, \quad
  \texttt{\{lisho,wutlin,migon,djiang\}@microsoft.com} \\
  \texttt{jpei@cs.sfu.ca}, \quad 
  \texttt{fuyan@uestc.edu.cn} \\
}

\date{}

\begin{document}
\maketitle
\begin{abstract}
In natural language processing (NLP) tasks, slow inference speed and huge footprints in GPU usage remain the bottleneck of applying pre-trained deep models in production. As a popular method for model compression, knowledge distillation transfers knowledge from one or multiple large (teacher) models to a small (student) model. When multiple teacher models are available in distillation, the state-of-the-art methods assign a fixed weight to a teacher model in the whole distillation. Furthermore, most of the existing methods allocate an equal weight to every teacher model.  In this paper, we observe that, due to the complexity of training examples and the differences in student model capability, learning differentially from teacher models can lead to better performance of student models distilled. We systematically develop a reinforced method to dynamically assign weights to teacher models for different training instances and optimize the performance of student model. Our extensive experimental results on several NLP tasks clearly verify the feasibility and effectiveness of our approach.

\end{abstract}

\section{Introduction}\label{sec:intro}

Deep pre-trained models, such as BERT~\cite{devlin2018bert}, XLNet~\cite{yang2019xlnet}, RoBERTa~\cite{liu2019roberta} and ALBERT~\cite{lan2019albert}, have proved effective on many NLP tasks by establishing record-breaking state-of-the-art results. However, due to huge amounts of model parameters, typically at the magnitude of hundreds of millions or even billions, the bottleneck for applying those pre-trained models in production is the slow inference speed and the huge footprints in using GPUs. To save the computation cost and speed up the inference process, {\em knowledge distillation} (KD)~\cite{hinton2015distilling} as an effective approach to compress large models into smaller ones stands out and becomes the de facto best choice among other alternatives, such as pruning~\cite{han2015compressing, han2015learning} and quantization~\cite{gong2014compressing}. 

The knowledge distillation approach is based on a teacher-student learning paradigm, where a teacher model, which is often large, is taken and the output of the teacher model is integrated as a soft target in the loss function to train a student model, which is often small. The teacher-student learning paradigm demonstrates excellent performance on various NLP tasks~\cite{kim2016sequence,tang2019distilling}. Knowledge distillation methods start from a single teacher. Some recent methods employ multiple teachers and show great promises in further boosting student model performance effectively.

\begin{table}[t]
\centering
\small
\begin{tabular}{llcc}  
\toprule
\multirow{2}{*}{\textbf{Teacher}} &   \multirow{2}{*}{\textbf{Student}}  & \textbf{MRPC}     & \textbf{MNLI-mm} \\
\cmidrule(l){3-4} 
&  & Acc  & Acc \\
\midrule
BERT-Base  & N / A          & 83.3  & 83.8 \\
RoBERTa-Base & N / A          &  88.5  & 86.8 \\
\midrule
BERT-Base  &   BERT$_3$     & 74.0 & 75.3  \\
RoBERTa-Base &   BERT$_3$     & 73.0  &  71.9 \\
\bottomrule
\end{tabular}
\caption{RoBERTa-Base performs better than BERT-Base. However, the student model distilled from BERT-Base, the weaker model, performs better than the same student model distilled from RoBERTa-Base, the stronger teacher model.}
\vspace{-1pt}
\label{tab:finding}
\end{table}

Most of the existing methods using multiple teachers simply assign an equal weight to all teacher models during the whole distillation process. The uniform distribution of weights among all teachers keeps the coordination, management and implementation of the multi-teacher framework simple.  At the same time, the indifference to strengths and weaknesses of various teacher models leaves a huge untouched space for better knowledge distillation.  To make a multi-teacher approach work well, teacher models have to be diverse.  Exploiting the diverse strengths of various teacher models can bring in huge advantages.

\emph{Individual teacher models may perform differently on various instances}. 
Different models may vary in hypothesis space, optimization strategy, parameter initialization and many other factors, which result in different performance among different cases.  Ideally, we want to assign different weights to different teacher models for different training instances according to their performance in individual cases.

Differentiating among teacher models is far from trivial.  Surprisingly, \emph{a stronger teacher model may not necessarily lead to a better student model}. As shown in Table~\ref{tab:finding} (Sun {\em{et  al.}}~\shortcite{sun2019patient}), the RoBERTa-Base model performs better than the BERT-Base model on the MRPC and MNLI-mm tasks. However, the student model using three-layer transformer BERT distilled from the weaker teacher model performs better on the same tasks than the same student model distilled from the stronger teacher model. One possible reason is that the effectiveness of distillation may be bounded by the capability of the student model. A simple student model with fewer parameters may not be able to approximate a very complex teacher model, since the complex teacher model may capture finer-grained patterns in data and cause the student model to overfit in some parts of the data and under some other parts. To achieve good distillation, we have to choose teacher models matching capacities of student models.

Based on the above insights, in this paper, we systematically study how to coordinate teacher models and student models in knowledge distillation.  Specifically, we investigate how to assign appropriate weights to different teacher models on various training samples. To the best of our knowledge, we are the first to treat teacher models deferentially at instance level in knowledge distillation.

We formulate the teacher model selection problem under a reinforcement learning framework: the decision is made based on the characteristics of training examples and the outputs of teacher models, while the policy is learned towards maximizing the student performance as the return. 

To verify the effectiveness of our approach in NLP, we conduct extensive experiments on several important tasks from the GLUE benchmark~\cite{wang2019glue}, including sentiment analysis, paraphrase similarity matching and natural language inference.  Our experimental results clearly show that our reinforced multi-teacher selection strategy boosts the performance of student models substantially.  Our method is not only principled but also practical in major NLP tasks.

Our contributions are twofold.  First, this is the first systematic study developing sample instance-based weighting for multiple teacher models. This is a concrete and novel contribution to knowledge distillation and machine learning. Our idea is general and can be used in many applications of knowledge distillation.  Second, we apply our novel idea of reinforced multi-teacher selection to a series of important NLP tasks.
This is a novel contribution to the NLP domain.  


\section{Related Work}
\label{sec:related-work}

To achieve model compression, \citet{hinton2015distilling} propose a knowledge distillation (KD) approach based on a teacher-student framework, which substantially extends the method by \citet{bucilu2006model}. The KD approach has been widely adopted in many applications. For example, \citet{kim2016sequence} demonstrate that standard knowledge distillation is effective for neural machine translation. Recent studies~\cite{tang2019distilling,sun2019patient,jiao2019tinybert} distill knowledge from BERT~\cite{devlin2018bert} into small student models and achieve competent results. All those methods distill knowledge from one teacher model.

To improve the performance of student models that employ deep neural networks, some recent studies leverage multiple teacher models in knowledge distillation. \citet{chebotar2016distilling} simply leverage the weighted average of teacher models to distill student model, where weights are hyper-parameters and fixed during training. \citet{fukuda2017efficient} examine two strategies to leverage labels from multiple teacher models in training student models. The first strategy updates the parameters of the student model by combining the soft labels from the teacher models with fixed weights. The other strategy randomly selects one teacher model at the mini-batch level to provide soft target labels for training student models. \citet{wu2019multi} also assign a fixed weight to each teacher model and use the weighted average of the probability distributions from multiple teacher models to train a student model.  \citet{yang2019model} propose a two-stage multi-teacher KD method, which first pre-trains a student model for the Q\&A distillation task and then fine-tunes the pre-trained student model with multi-teacher distillation on downstream tasks. Teacher models are assigned with equal weights during distillation.

All previous knowledge distillation methods using multi-teacher models fix the same weight for a teacher model on all training examples.  In this paper, we learn a policy to dynamically assign weights to teacher models based on individual examples.

\section{Our Approach}
\label{sec:approach}

In this section, we first recall the preliminaries of knowledge distillation and teacher models/student models.  Then, we present our reinforced teacher selection method.  Last, we introduce the model training algorithm.

\subsection{Preliminaries}
\label{sec:vanilla}

In general, the knowledge distillation approach~\cite{bucilu2006model,hinton2015distilling} uses the soft output (logits) of one or multiple large models as the knowledge and transfers the knowledge to a small student model. In this paper, we mainly target at NLP tasks and thus use the state-of-the-art NLP models as examples to illustrate our approach. Our approach in general can be applied to all knowledge distillation tasks using teacher/student models.
 
Given a NLP task, let $\mathcal{D}=\{(\mathbf{x}_i, \mathbf{y}_i)\}^N_{i=1}$ be the training set with $N$ training examples, where $\mathbf{x}_i$ is the $i$-th input, such as a single sentence or a pair of sentences, and $\mathbf{y}_i$ is the corresponding ground truth label.  Without loss of generality, we assume that a class label $c$ is an integer between $1$ and $C$, where $C$ is the number of classes in the data set.

We assume $K$ teacher models.  For example, we can fine-tune the pre-trained models with 12-layer transformers, such as BERT~\cite{devlin2018bert}, XLNet~\cite{yang2019xlnet}, RoBERTa~\cite{liu2019roberta} or ALBERT~\cite{lan2019albert}, on the training data as possible teacher models. Denote by $\mathcal{M}_{k}$ ($1\le k\le K$) the $k$-th teacher model, and by $\Theta^t_k$ the set of model parameters of $\mathcal{M}_{k}$. The output of teacher model $\mathcal{M}_{k}$ for a given input $\mathbf{x}_i$ is written as $\mathbf{\tilde{y}}_{i,k}^t =\langle \tilde{y}_{i,k,1}^t, \ldots, \tilde{y}_{i,k,C}^t \rangle$, where $\tilde{y}_{i,k,c}^t =P^t(\mathbf{y}_i=c|\mathbf{x}_i; \Theta^t)$ is the probability of $\mathbf{x}_i$ belonging to class $c$ $(1 \leq c \leq C)$ computed by model $\mathcal{M}_{k}$.  

For student models, we consider transformer models with fewer layers, such as those with only $3$ or $6$ layers. To train a student model, the distillation loss is defined as the cross-entropy loss between the predictions of the teacher model and that of the student model, that is,
\begin{equation}\small
    \label{eq:dl}
        \mathcal{L}_{DL}=-\frac{1}{K}\sum_{i, k, c} \left [\tilde{y}_{i, k, c}^t \cdot \mathrm{log}P^s(\mathbf{y}_i=c|\mathbf{x}_i;\Theta^s)\right ]
\end{equation}
where $\Theta^s$ is the set of parameters for the student model and $P^s(\mathbf{y}_i=c|\mathbf{x}_i;\Theta^s)$ is the probability of example $\mathbf{x}_i$ belonging to class $c$ computed by the student model. 

Besides the distillation loss, a ground-truth loss is also introduced in the training of student model.
\begin{equation}\small
    \label{eq:ce}
    \begin{aligned}
        \mathcal{L}_{CE} = - \sum_{i, c}  \mathbbm{1} \left [  \mathbf{y}_i=c \right ] \cdot \mathrm{log} P^s(\mathbf{y}_i=c|\mathbf{x}_i;\Theta^s)
    \end{aligned}    
\end{equation}
where $\mathbbm{1} \left [  \mathbf{y}_i=c \right ]$ is the one-hot vector from the ground truth label.  The final objective of KD is
\begin{equation}
    \label{eq:kd}
    \begin{aligned}
        \mathcal{L}_{KD} = \alpha{\mathcal{L}_{DL}} + (1 - \alpha) \mathcal{L}_{CE}
    \end{aligned}    
\end{equation}
where $\alpha$ is a hyper-parameter to balance between the soft teacher label $\mathcal{L}_{DL}$ and the hard ground-truth label $\mathcal{L}_{CE}$.

\subsection{RL based Teacher Selector (TS)}

The original KD methods~\cite{bucilu2006model,hinton2015distilling} assumes only one single teacher model. Although several recent studies employ multiple teacher models in distillation, as reviewed in Related Work, those methods assign a fixed weight to each teacher model. Inspired by the insights discussed in Introduction, we propose to dynamically assign weights to teacher models at instance level using a reinforced approach~\cite{sutton1992reinforcement}. The weights are implemented as the sampling probabilities calculated by the reinforcement learning policy function, as to be explained in Equation~\ref{eq:policy}.

\begin{figure}[t]
    \centering
    \includegraphics[trim={1.5cm 5.7cm 14cm 7cm}, clip, scale=0.56] {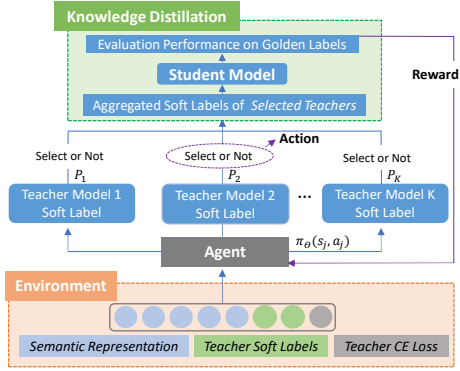}
    \caption{Overview of RL-KD, Reinforced Multi-Teacher Selection for Knowledge Distillation.}
    \label{fig:overview}
\end{figure} 

Figure~\ref{fig:overview} shows the overview of our Reinforcement Learning based Knowledge Distillation approach (RL-KD for short). In each iteration, an agent interacts with the environment and receives the representation of a training instance. The agent then applies the teacher selection policy and samples the teachers to participate in the knowledge distillation step. The outputs of the sampled teachers are integrated into a soft label to train the student model. After an episode of training examples, the performance of the trained student model is used as the reward to update the policy parameters. This process iterates on episodes until the performance of the student model converges.

In general, a reinforcement learning approach involves elements in the form of (\emph{state, action, reward}). The elements in our method are as follows.

\paragraph{State} 
Our reinforcement learning method maintains a series of environment states $s_1, s_2, \ldots$ that summarize the characteristics of the input instances as well as the candidate teacher models, so that judicious decisions can be made accordingly. We design a state $s_j$ as a vector of real-values $\mathbf{F}(s_j)$, which includes the concatenation of three features.

The first feature is a {\bf vector representation} $\mathcal{R}(\mathbf{x}_i) \in \mathbb{R}^d$  of an input instance $\mathbf{x}_i$. In general, any sentence-encoding network can be applied here to embed a sentence in a NLP task into a vector.  In this paper, we use the embedding for the [CLS] token in the last layer of the pre-trained BERT-Base model~\cite{devlin2018bert}.

The representation model $\mathcal{R}$ here is different from a BERT-Base teacher model $\mathcal{M}_k$. $\mathcal{R}$ is a pre-trained BERT-Based model aiming to effectively represent the content of input instances. A teacher model $\mathcal{M}_k$ is fine-tuned for a specific task, such as predicting the class label distribution given an input instance. 

The second feature is the {\bf probability vector} $\{P^t_k(\mathbf{y}_i=c|\mathbf{x}_i; \Theta^t_k)\}_{c=1}^C$ on all classes $1, \ldots, C$ predicted by the teacher model $\mathcal{M}_k$ with parameters $\Theta^{t}_k$ on the input instance $\mathbf{x}_i$.  In practice, the probability is often derived from a softmax function on the hidden representation of the input, that is, 
\begin{equation}\label{eq:class_prob}\small
    P^t_k(\mathbf{y}_i=c|\mathbf{x}_i; \Theta^t_k)=\left [ \mathrm{softmax}(\mathbf{W_k}\cdot \mathcal{M}_{k}(\mathbf{x_i)}) \right ] _c
\end{equation}
where $\left [ \cdot \right ] _c$ refers to the $c$-th element in a vector, $\mathbf{W_k} \in \mathbb{R}^{C \times d}$ is a trainable weight matrix, and $\mathcal{M}_k(\mathbf{x}_i)$ is the embedding of the [CLS] token in the last layer of $\mathcal{M}_k$.

The third feature is the {\bf cross entropy loss} $\mathcal{L}_{k}(\mathbf{x}_i)$ on the input instance $\mathbf{x}_i$ of the teacher model $\mathcal{M}_k$ with parameters $\Theta^{t}_k$, which is the ground-truth loss of teacher model for $\mathbf{x}_i$.
\begin{equation}\small
    \label{eq:ce_teacher}
    \begin{aligned}
        \mathcal{L}_{k}(\mathbf{x}_i) = - \sum_{c}  \mathbbm{1} \left [  \mathbf{y}_i=c \right ] \cdot \mathrm{log} P^t_k(\mathbf{y}_i=c|\mathbf{x}_i;\Theta^t_k)
    \end{aligned}    
\end{equation}

For the second and third features, we also concatenate the predictions of other teacher models as state features for the policy agent to better select teachers. 
    
\paragraph{Action} Each teacher model $\mathcal{M}_k$ is associated with one agent. An agent chooses between two possible actions, selecting the teacher model or not for the current instance. A policy function $\pi _\theta(s_j, a_j)$ determines the distribution over the states, from which the action value of $a_j \in \{0, 1\}$ is sampled. Denote by $\theta$ the trainable parameters in the policy function. In this paper, we adopt a simple logistic function as the policy model. 
\begin{equation}
\label{eq:policy}
\small
\begin{aligned}
        \pi _\theta(s_j, a_j) =& P_{\theta}(a_j|s_j) \\
        =& a_j\sigma(\mathbf{AF}(s_j)+\mathbf{b}) \\
        & + (1 - a_j)(1 - \sigma(\mathbf{AF}(s_j)+\mathbf{b}))
\end{aligned}
\end{equation}
where $\mathbf{F}(s_j)\in \mathbb{R}^{d+(C+1)*K}$ is the state vector and $\sigma(\cdot)$ is the sigmoid function with trainable parameter $\theta = \{\mathbf{A} \in \mathbb{R}^{d+(C+1)*K},  \mathbf{b} \in \mathbb{R}^{1}\}$.
The probabilities defined by the above policy function are the weights assigned to the teacher models.

\paragraph{Reward} The reward function is correlated with the performance of the student model trained from the distillation of the selected teacher models. We define an episode $\xi_b$ as one batch of training instances, that is, $\mathcal{D}_b=\{\mathbf{x}_i$, $\mathbf{x}_{i+1}$, \ldots, $\mathbf{x}_{i+m-1}\}$, where $b$ denotes the batch ID and $m$ is the batch size. For each instance $\mathbf{x}_j$ ($i\le j\le i+m-1$), we construct the state vector $\mathbf{s}_{j_k}$ for each teacher model $\mathcal{M}_k$ and sample the actions $a_{j_k}$ according to policy $\pi_{\theta}(s_{j_k}, a_{j_k})$ (Equation~\ref{eq:policy}). For all $a_{j_k}$ sampled, we integrate the average of the loss $\mathcal{L}_{DL}$ (Equation~\ref{eq:dl}) into the loss $\mathcal{L}_{KD}$ (Equation~\ref{eq:kd}) to train the student model.

We propose three reward calculation methods. The first is the minus of ground truth loss $\mathcal{L}_{CE}$ (Equation~\ref{eq:ce}) of the student model. The second reward function combines both the minus of ground-truth loss and the distillation loss to measure the student model. Last, in order to encourage better generalization, we further take the accuracy metric of student model on the development set\footnote{\scriptsize Following the common practice in building machine learning models, we assume that data used are divided into training data, validation/development data, and testing data.  Development set is used to tune hyper-parameters.} $\mathcal{D}'$ as the reward.  Other metrics generated on the development set could also be applicable. In summary, we have
\begin{equation}
\label{eq:reward}
    \begin{split}
            reward_1 = & -\mathcal{L}_{CE} \\
            reward_2 = & -\mathcal{L}_{CE} - \mathcal{L}_{DL} \\
            reward_3 = & \gamma * (-\mathcal{L}_{CE} - \mathcal{L}_{DL}) \\
            & + (1 - \gamma) * Accuracy \, on \, \mathcal{D}' \\
    \end{split}
\end{equation}
where $\gamma$ is a hyper-parameter to balance the reward from the training set and the development set.
Please note that the reward is not given immediately after each step is taken. Instead, it is delayed until the training of the whole batch is completed. 


 
\paragraph{Optimization} We follow the standard policy gradient method to optimize the parameters, that is,
\begin{equation}
\label{eq:op}
    \theta \leftarrow \theta + \beta \sum_j {r \sum_k \bigtriangledown_\theta\mathrm{\pi}_\theta{(s_{j_k}, a_{j_k})}}
\end{equation}
where $r$ is defined by Equation~\ref{eq:reward} and $\beta$ is the learning rate.


\subsection{Model Training} 

\begin{algorithm}[t]
\small
1. Pre-train the student model $\Theta^s_0$ using knowledge distillation from the average of all teacher models by maximizing $\mathcal{L}_{KD} = \alpha{\mathcal{L}_{DL}} + (1 - \alpha) \mathcal{L}_{CE}$.

2. Pre-train the TS policy $\theta_0$ by calculating the  return under $\Theta^s_0$  with all teacher models selected.

3. Run Algorithm~\ref{alg:rl-ts} to iteratively train KD and TS in turn until convergence.
\caption{Overall Training Procedure}
\label{alg:complete}
\end{algorithm}

Algorithm~\ref{alg:complete} shows the overall RL-KD approach. At the beginning, we initialize the student model $\Theta^s_0$ by employing all teacher models in the knowledge distillation, that is, $a_{j_k}$ is set to 1 for all $\mathcal{M}_k$ and for all instances $\mathbf{x}_i$. We then initialize parameter $\theta$ for the policy function using the same setting under $\Theta^s_0$. After initialization, we iteratively conduct knowledge distillation and teacher selection in turn. 

\begin{algorithm}[t]
\small
\KwIn{Epoch number $L$. Training data $\mathcal{D}=\{\mathcal{D}_1, \mathcal{D}_2,...,\mathcal{D}_M\}$. A KD model and a TS model initialized as $\Theta^{s}=\Theta^{s}_0$ and $\theta=\theta_0$.} 

\For {epoch $l = 1$ to $L$}   
{
	Shuffle $\mathcal{D}$ to obtain a new training sequence. \\
	\For {each batch $\mathcal{D}_b \in \mathcal{D}$} 
	{
	    TS sample actions for each instance $\mathbf{x}_i \in \mathcal{D}_b$ with $\theta$ to get selected teachers $\mathcal{K}$ by: $a_{j}\sim \pi_{\theta}(s_j, a_j)$. \\ 
		Stored $(a_{j}, s_j)$ to the episode history $\mathcal{H}$.\\
		Compute the average of soft labels of the selected teachers:\\
		\quad $\frac{1}{|\mathcal{K}|}\sum_{k\in\mathcal{K}}{P^t_k(\mathbf{y}_i=c|\mathbf{x}_i; \Theta^t_k)}$ \\
		Update the parameter $\Theta^s$ of KD by: \\
		\quad $\mathcal{L}_{KD} = \alpha{\mathcal{L}_{DL}} + (1 - \alpha) \mathcal{L}_{CE}$\\
	}
	\For {each $(a_{j}, s_j) \in \mathcal{H}$}
	{
		Compute delayed reward following Equation~\ref{eq:reward}. \\
		Update the parameter $\theta$ of TS following Equation~\ref{eq:op}. \\
	}
}
\caption{Joint Training of TS and KD}
\label{alg:rl-ts}
\end{algorithm}

As described in Algorithm~\ref{alg:rl-ts}, in the KD process, we fix the teacher selection policy $\theta$ and learn the student model $\Theta^s$. In the TS process, we fix $\Theta^s$ to calculate the return and optimize the teacher selection policy $\theta$. The iteration continues for $L$ epochs. For more implementation details, please refer to Section ``Implementation Details''.

\section{Experiments}
\label{sec:experiment}

In this section, we first describe our experimental setup, and then introduce the baseline methods and implementation details of our method.  Last, we report the experimental results.

\begin{table}[t]
\small
\centering
\scalebox{0.97}{\begin{tabular}{lccclccc}  
\toprule
Dataset & \#Train    & \#Dev  & \#Test \\
\midrule
RTE       & 2,490   & 277    & 3,000      \\
MRPC      & 3,668   & 408    & 1,725      \\ 
SST-2     & 67,349  & 872    & 1,821      \\
QNLI      & 104,743 & 5,463   & 5,463     \\
MNLI-mm   & 392,702 & 9,832   & 9,847     \\
MNLI-m    & 392,702 & 9,815   & 9,796     \\
QQP       & 363,849 & 40,430  & 390,965  \\
\bottomrule
\end{tabular}
}
\caption{Statistics of the datasets for experiment.}
\label{tab:stat-dataset}
\end{table}

\subsection{Data Sets and Evaluation Metric}
Following Patient KD~\cite{sun2019patient}, we evaluate our proposed approach on three different NLP tasks from the GLUE benchmark~\cite{wang2019glue}, namely Sentiment Classification (SC), Paraphrase Similarity Matching (PSM) and Natural Language Inference (NLI).
The statistics of the data sets are shown in Table~\ref{tab:stat-dataset}. We use prediction {\em accuracy} as the metric in evaluation.

For SC, we experiment on Stanford Sentiment Treebank (SST-2)~\cite{socher2013recursive}. The target is to predict the sentiment of a given sentence.

For PSM, we use Microsoft Research Paragraph Corpus (MRPC)~\cite{dolan2005automatically} and Quora Question Pairs (QQP)~\cite{wang2019glue}. The goal on both data sets is to decide whether two sentences of questions are semantically equivalent.

For NLI, we evaluate on Multi-Genre Natural Language Inference (MNLI)~\cite{williams2017broad}, Question-Answering Natural Language Inference (QNLI) and Recognizing Textual Entailment (RTE). MNLI is a corpus of sentence pairs extracted from multiple domains, which are further divided into two splits, in-domain (MNLI-m) and cross-domain (MNLI-mm), to evaluate the model generality. QNLI is a data set converted from a question-answering data set SQuAD~\cite{rajpurkar2016squad}, and is designed to predict whether the context sentence contains an answer to the question. RTE is based on a series of textual entailment challenges from General Language Understanding Evaluation (GLUE)~\cite{wang2018glue}.

\subsection{Baseline Models}

We fine-tune the pre-trained models BERT$_{12}$, RoBERTa$_{12}$, ALBERT$_{12}$ and XLNet$_{12}$ on task specific training data to get four candidate teacher models, where the subscript 12 means 12 layers of transformers. Similar to Patient KD~\cite{sun2019patient}, we consider the BERT models with 3 layers and 6 layers of transformers as two student models, denoted by BERT$_{3}$ and BERT$_{6}$, respectively.

To validate the effectiveness of our proposed method, we compare with various baselines reviewed in Related work. 

{\textbf{Single Teacher}} KD~\cite{hinton2015distilling} is applied to train student models from BERT$_{12}$, RoBERTa$_{12}$, ALBERT$_{12}$ and XLNet$_{12}$ individually, which is referred to as \emph{Vanilla KD (V-KD)} in the rest of this section. 

{\textbf{U-Ensemble Teacher}} is our implementation of the equal weight method by~\citet{yang2019model}. Every teacher model is assigned an equal weight in KD, and the student model learns from an aggregated distribution by averaging the outputs of all teacher models. 

{\textbf{Rand-Single-Ensemble Teacher}} uses the strategy of~\citet{fukuda2017efficient} to randomly select one teacher from the teacher model candidates at mini-batch level to provide soft-targets for training student model.
    
{\textbf{W-Ensemble Teacher}} is a weighted assemble method following~\citet{chebotar2016distilling, fukuda2017efficient, wu2019multi}. We assign a different weight to each teacher model. The weights are fixed during the whole distillation.

Besides these existing baseline methods, we further propose two strong baselines for comparison. 
    
{\textbf{LR-Ensemble Teacher}} uses Logistic Regression (LR) model to model the best weights for each teacher candidate, instead of setting weights using heuristics like U-Ensemble/W-Ensemble. Specifically, let the aggregated distribution $P(\mathbf{y}_i=c|\mathbf{x}_i)=\sum_k w_k P^t_k(\mathbf{y}_i=c|\mathbf{x}_i; \Theta^t_k)$. We learn $w_k$ by maximizing the performance of the weighted ensemble teacher model. Since the training of the best weights can be conducted on either the training set or the development set, we denote by \textbf{LR-Train-Ensemble} and \textbf{LR-Dev-Ensemble} the corresponding LR models, respectively.

All the above baseline methods either use a single teacher model or assign fixed weights to ensemble multiple teacher models. As the last baseline, \textbf{Best-Single-Ensemble Teacher} considers an instance-level teacher model selection method, where for each instance in the training set (where ground-truth is available), the best performing single teacher model, that is, the one achieving the lowest cross entropy loss, is selected to train student model. We can treat this model as the upper bound of selecting best teacher model. Please note that this model cannot be evaluated on any test set since we are not supposed to use the ground-truth labels there for tuning models. 

\begin{table*}[t]
\small
\centering
\begin{tabular}{m{30mm}cccccccc}  
\toprule
 \multirow{1}{*}{\textbf{Teacher Model}} & \textbf{QQP} & \textbf{MRPC} & \textbf{MNLI-mm}  & \textbf{RTE} & \textbf{MNLI-m}  & \textbf{QNL}I  & \textbf{SST-2} & \multirow{1}{*}{\textbf{AVG.}} \\
\midrule
BERT$_{12}$       &   90.9    &    83.3    &    83.8    &    63.9    &    83.6    &    91.3    &    92.1  & 84.1 \\
RoBERTa$_{12}$      &       91.3    &    88.5    &    86.8    &    70.8    &    86.5    &    91.9    &    93.7 & 87.1 \\
XLNet$_{12}$       &     91.0    &    87.5    &    85.8    &    71.8    &    85.3    &    91.5    &    93.8 & 86.7 \\
ALBert$_{12}$      &       90.6     &    88.7    &    84.3    &    74.0    &    83.9    &    91.6    &    91.2 & 86.3\\
\midrule
U-Ensemble      &   \textbf{92.3}    &    89.2    &    \textbf{87.5}   &    72.6    &    87.3    &    \textbf{93.0}    &    93.7 & 87.9 \\
W-Ensemble       &   \textbf{92.3}    &    89.5    &    \textbf{87.5}    &    72.9    &   \textbf{87.4}    &    \textbf{93.0}    &    93.6 & 88.0\\
Rand-Single-Ensemble & 90.7 &	88.5 &	85.4 &	72.6 &	84.4 &	91.3 &	92.7 &	86.5 \\
LR-Train-Ensemble   &    92.1    &    89.5    &    86.7    &    71.8    &    86.5    &    92.5    &    92.9 & 87.4\\
LR-Dev-Ensemble    &    \textbf{92.3}    &    \textbf{90.0}    &    87.3    &    \textbf{74.4}    &    87.1    &    \textbf{93.0}    &    \textbf{94.0} & \textbf{88.3} \\
\bottomrule
\end{tabular}
\caption{Performance of various teacher models on test sets, including both individual models and ensemble models by various ensemble strategies.}
\label{tab:teachers-ensemble}
\end{table*}

\subsection{Implementation Details}
\label{sec:imp}
The training code is built on top of the code repository of Patient KD~\cite{sun2019patient}\footnote{\scriptsize \url{https://github.com/intersun/PKD-for-BERT-Model-Compression}.}. All tasks in our experiments can be treated as classification problems where the input is either a single sentence or a pair sentences. For the tasks whose inputs are individual sentences, the model input has the form of {\tt [CLS]} $sentence_1$ {\tt [SEP]}. For tasks whose input is sentence pairs, the input form is {\tt [CLS]} $sentence_1$ {\tt [SEP]} $sentence_2$ {\tt[SEP]}. 

To fine-tune the teacher models, we adopt the open-sourced pre-trained weights for BERT$_{12}$, RoBERTa$_{12}$, ALBERT$_{12}$ and XLNet$_{12}$ as initialization. The learning rate is set to \{1e-5, 2e-5, 5e-5\}. The batch size is set to $32$. The maximum sequence length is set to $128$. The number of epochs is set to $4$. The best model is selected according to the accuracies on the development set.

The student models, BERT$_3$ and BERT$_6$, are initialized by the bottom 3 and 6 layers of BERT-Base\footnote{\scriptsize \url{https://github.com/google-research/bert}.}, respectively. 
Meanwhile, we set the batch size to $32$, the number of epochs to 4, the maximum length of sequence to $128$, the learning rate to \{1e-5, 2e-5, 5e-5\}, the distillation temperature $T$ to \{5, 10, 20\}, and the loss equilibrium coefficient $\alpha$ to \{0.2, 0.5, 0.7\}. We choose the best model based on the performance on the development set. The $\gamma$ in the experiments ranges from \{0.3, 0.5, 0.7, 0.9\}, which is selected based on development set performance. 

For the teacher model selector, our policy function is a simple logistic regression model. After feeding the input sequence encoded by the original pre-trained BERT-Base model into the teacher model selector, we adopt a standard Monte-Carlo based policy gradient method~\cite{williams1992simple} to optimize the parameters of the policy model.

{\textbf{KD Pretraining}} Our KD model is initialized using the pre-trained BERT model weights. Then, we use the distillation task in question to pre-train the KD model to learn from the average ensemble teacher model (i.e., average of all teacher model predictions). We set the batch size to 32 and max epochs to 4. The other hyper-parameters are kept the same as the normal KD training mentioned above.  

{\textbf{TS Pretraining}} Pretraining is recommended by many reinforcement learning methods~\cite{bahdanau2016actor,feng2018reinforcement}. For TS pretraining, to avoid the instability of initial student model and speed up the training process, we leverage the performance of the average ensemble of the selected teacher models as the reward to train the TS Model.

{\textbf{Iterative Training}}   After the pretraining stage, the KD and TS models are jointly trained alternatively in a batch-wise way. At batch \#1, KD is trained while keeping TS fixed. At batch \#2, TS is trained while keeping KD fixed, so on and so forth. If TS selects no teacher, this instance will not receive any reward from KD.


\begin{table*}[t]
\small
\centering
\begin{tabular}{lccccccccc} 
\toprule
\midrule
\textbf{Teacher}   &  \textbf{Student}  &  \textbf{Strategy}  &   \textbf{QQP}   &  \textbf{MRPC}   &  \textbf{MNLI-(mm/m)}   &  \textbf{RTE}  &  \textbf{QNLI}   &  \textbf{SST-2} & \textbf{AVG.}\\
 \midrule
-   &   BERT${_3}$   &   FT   &   88.3   &   72.1   &   74.6 /  74.8   &   60.7      &   83.3   &   85.9  & 77.1\\
BERT$_{12}$   &   BERT${_3}$   &   V-KD   &   88.4   &   74.0   &   75.3 / 75.6   &   56.7      &   83.7   &   87.5 & 77.3 \\
Robert$_{12}$   &   BERT${_3}$   &   V-KD   &   85.3   &   73.0   &   71.9 /  71.7   &   55.2    &   81.9   &   86.0 & 75.0 \\
XLNet$_{12}$   &   BERT${_3}$   &   V-KD   &   85.4   &   74.3   &   72.0  / 71.6  &   55.2      &   81.4   &   87.4 & 75.3 \\
ALBERT$_{12}$   &   BERT${_3}$   &   V-KD   &   88.4   &   74.0   &   76.4 / 75.6  &   56.7     &   83.7   &   87.5 & 77.5 \\
Rand-Single-Ensemble & BERT${_3}$   &   V-KD     & 87.3	& 70.6 &	75.0 / 74.5 &	51.6  &	83.7 &	86.0 &	75.5 \\
W-Ensemble &  BERT${_3}$   &   V-KD   & 85.3 &	74.8 &	72.2 / 72.0 &	56.7	&	82.4	& 87.5 &	75.8 \\
LR-Dev-Ensemble   &   BERT${_3}$   &   V-KD   &   88.6   &   71.8   &   75.4 /  75.4  &   54.9    &   84.2   &   86.8 & 76.7 \\
Best-Single-Ensemble   &   BERT${_3}$   &   V-KD   &   88.6   &   \textbf{77.2}  &   75.1 / 74.7  &   56.0     &   84.1   &   87.0 & 77.5\\
\hdashline
Our Method ($reward_1$)   &   BERT${_3}$   &   RL-KD   &   \textbf{89.1}   &   76.0  &  76.9 / \textbf{76.8}  &  61.4   &   \textbf{85.4}   &   \textbf{89.1} & 79.2 \\
Our Method ($reward_2$)    &   BERT${_3}$   &   RL-KD   &   \textbf{89.1}   &   76.2   &   \textbf{77.4} / 76.3  &   63.5  &   84.8  &   88.5 & 79.4 \\
Our Method ($reward_3$)   &   BERT${_3}$   &   RL-KD   &   89.0   &   76.7   &   76.7 / 75.7  &   \textbf{64.6}  &   85.3  &   \textbf{89.1} & \textbf{79.6} \\
\midrule
 - &   BERT${_6}$   &  FT   &   90.0   &   82.1   &   80.7 / 80.3 &   63.2    &   87.1   &   90.1 & 81.9 \\
BERT$_{12}$   &  BERT${_6}$   &   V-KD   &   90.1   &   82.6   &   80.5 / 81.2  &   63.5    &   88.4   &   90.8 & 82.4\\
Robert$_{12}$   &   BERT${_6}$   &   V-KD   &   87.2   &   80.2   &   78.0 /  77.0   &   57.4   &   84.5   &   90.5 & 79.3 \\
XLNet$_{12}$   &   BERT${_6}$   &   V-KD   &   87.4   &   80.6   &   77.5 / 76.7  &   62.1   &   84.1   &   91.1 & 79.9 \\
ALBERT$_{12}$   &   BERT${_6}$   &   V-KD   &   90.2   &   79.7   &   82.3 / 81.1   &   65.3  &   88.3   &   89.9 & 82.4 \\
Rand-Single-Ensemble &  BERT${_6}$   &   V-KD   &  89.7 &	77.7 & 81.0 / 80.7 & 61.7 &	87.6 & 90.6 & 81.3\\
W-Ensemble &  BERT${_6}$   &   V-KD   &  87.3 &	81.1 & 77.6 / 77.2  &	62.1	& 84.8  &	90.6 & 80.1  \\
LR-Dev-Ensemble   &   BERT${_6}$   &   V-KD   &   90.3   &   80.6   &   81.3 / 81.1  &   64.6    &   88.3   &   90.8 & 82.4 \\
Best-Single-Ensemble   &   BERT${_6}$   &   V-KD   &   90.1   &   80.4   &   80.7 / 80.5   &   66.1  &   87.1   &   90.3 & 82.2 \\
\hdashline
Our Method ($reward_1$)   &   BERT${_6}$   &   RL-KD   &   \textbf{90.7}   &   82.8   &   \textbf{82.3} / 82.0  &   67.1   &   88.7   &  91.7 & 83.6 \\
Our Method ($reward_2$)   &   BERT${_6}$   &   RL-KD   &   90.6   &   82.1   &   \textbf{82.3} / 82.1  &   67.2  &   88.8  &   91.4  & 83.5 \\
Our Method ($reward_3$)   &   BERT${_6}$   &   RL-KD   &   90.5   &   \textbf{83.3}   &   82.2 / 81.6  &   \textbf{68.2}  &   \textbf{88.8}  &   \textbf{92.3} &\textbf{83.8} \\
\bottomrule
\end{tabular}
\caption{Student model performance on test set. We compare three training strategies for student models: (1) \textbf{FT:} The student model is directly fine-tuned on task label data without any soft target labels from teacher models. (2) \textbf{V-KD:} vanilla KD from single teacher, teacher ensemble, or the best single teacher for each instance. (3) \textbf{RL-KD:}  our RL method to select from multiple teacher models at instance level.  For RL-KD, we consider all single model teachers (BERT$_{12}$, Robert$_{12}$, XLNet$_{12}$, ALBERT$_{12}$). For our method, we compare three different reward functions: (1) $reward_1$: we use the minus of ground-truth loss (CE) of student model as the reward function for teacher model selection. (2) $reward_2$: Besides the minus of ground-truth loss (CE), we also introduce the minus of knowledge distillation loss (DL) into the reward function.  (3) $reward_3$: Besides $reward_2$, we also take the Accuracy metric of student model on development set into account for better generalization.
}
\label{tab:experimental-res}
\end{table*}

\begin{table}[t]
\small
\centering
\begin{tabular}{lccc}  
\toprule
 & \textbf{RTE}                                     
& \textbf{Mean} & \textbf{Stdev} \\
W-Ensemble & 56.7, 57.0, 57.0, 57.8, 58.1  & 57.32 & 0.597  \\
LR-Ensemble & 	53.8, 56.0, 57.8, 58.1, 58.5 & 56.84 & 1.950 \\
RL-KD$\P\spadesuit$ &  61.7, 63.2, 63.9, 64.6, 64.6 & 63.60 & 1.083	 \\
\midrule
 & \textbf{MRPC} & \textbf{Mean} & \textbf{Stdev}	 \\
W-Ensemble & 71.6, 72.3, 72.5, 74.7, 74.8  & 73.18 &	1.472 \\
LR-Ensemble & 72.1, 72.1, 73.8, 74.3, 74.8  & 73.42 & 1.256 \\
RL-KD$\P\spadesuit$ & 74.7, 75.0, 75.5, 75.7, 76.7 & 75.52 &	0.688 \\
\bottomrule
\end{tabular}
\caption{Mean and Standard deviation values and statistically significance T-test (p-Value) across five different runs on RTE and MRPC. $\P$ and $\spadesuit$ denote statistically significant improvements over W-Ensemble and LR-Ensemble.}
\label{tab:variance-analysis}
\end{table}


\subsection{Experimental Results} \label{sec:results}

\paragraph{Results on Teacher Models}  We start with the performance of individual teacher models and ensemble teacher models shown in Table~\ref{tab:teachers-ensemble}. 
\emph{Individual teacher models perform differently in different tasks}. No one single model wins on all tasks. RoBERTa achieves better results than the other models in three tasks (MNLI-mm, MNLI-m, QNLI), ALBERT shows higher accuracies on tasks MRPC and RTE, while XLNet scores the highest on SST-2. This suggests that different models may learn different local optimum and bear different biases. 
    
Moreover, \emph{ensemble of teacher models can lead to better performance.}
All ensemble teacher models achieve better results than a single teacher model. This verifies that model ensemble is effective to mitigate the biases in various teacher models. By comparing different ensemble strategies, the weighted ensemble methods including W-Ensemble and LR-Ensembles outperform U-Ensemble. It indicates that smartly assigning weights to different models can lead to better results. Among all the ensemble models, LR-Dev-Ensemble achieves the best performance. Intuitively, training the weights based on the development set may help this model gain better generalization capability. Therefore, in the next experiments where we compare student models, we pick this best teacher model as the strongest baseline to represent ensemble methods. 

\paragraph{Results on Student Models}

The performance of student model is shown in Table~\ref{tab:experimental-res}. 
\emph{A stronger teacher model may not necessarily lead to a better student.} Table~\ref{tab:teachers-ensemble} shows that ensemble teacher models are better than single teacher models. However, the corresponding student models distilled from those stronger teacher models are not always stronger. For example, the BERT$_3$ student model distilled from ALBERT$_{12}$ is better than that from LR-Dev-Ensemble on MNLI-mm, MNLI-m and SST-2. The BERT$_6$ student model distilled from XLNET$_{12}$ is also better than or on par with that from LR-Dev-Ensemble on MRPC and SST-2. 

Even in the extreme case where Best-Single-Ensemble is used as the teacher model and we always choose the best performing teacher whose output is the closest to the ground-truth label, the corresponding student model does not always perform the best. Take BERT$_6$ as example. The student model from the stronger teacher, Best-Single-Ensemble, is inferior to that from LR-Dev-Ensemble on almost all data sets. This observation indeed motivates our RL-KD method, that is, the best teacher model may not necessarily performs the best in distillation. Otherwise, we just use the ground truth label as the teacher and do not need KD in the first place. 
     
\emph{RL-KD consistently outperforms the other methods}, including those strong baselines with ensemble teacher models. The student models (BERT$_3$ and BERT$_6$) trained by our proposed RL-KD method show consistently better results on most cases, while the performance of the student models distilled from different baseline KD methods varies on different data sets. This verifies that our proposed RL-KD method learns to adapt to the capability of the student models and dynamically selects the ``most suitable teachers'' in the training of student models. Among the three reward functions proposed, $reward_3$ performs the best. This is consistent with our intuition that this reward prefers the student that not only fits the training set well, but also is able to generalize to validation/development set by adding the acc on development set as reward. 

The only exception is on the MRPC set for BERT$_3$, where the size of training data for this task is relatively small (see Table~\ref{tab:stat-dataset}). For a weak student model such as BERT$_3$, when the training size is small, the reward computed on top of this model may not be robust, which is also indicated by the relatively lower accuracy on the MRPC sets. Consequently, the policy for teacher model selection may be compromised. As a comparison, for student model BERT$_6$, the model generalization ability is stronger. Therefore, even a small size of data can already yield a reasonable model, which in turn provides reliable reward as the feedback for policy training. That explains why student model BERT$_6$ achieves better performance than the other KD methods even on relatively small data sets.




\subsection{Variance Analysis}

We conduct 5 runs of models training and calculate the mean and standard deviation (Stdev) values. Besides, we also conduct a two-sided statistically significant t-test (p-Value) with threshold $0.05$ comparing baseline methods with our RL-KD method. The experimental results are listed in Table~\ref{tab:variance-analysis}. Results shows that (1) the variance of our approach is similar to baseline KD. (2) our method outperforms baselines with statistical significance.


\section{Conclusions}
\label{sec:conclusion}
In this paper, we tackle the problem of teacher model selection in knowledge distillation when multiple teacher models are available. We propose a novel RL-based approach, which dynamically assigns weights to teacher models at instance level to better adapt to the strengths of teacher models as well as the capability of student models. The extensive experiments on several NLP tasks verify the effectiveness of our proposed approach.  

\section*{Acknowledgments}

Jian Pei's research is supported in part by the NSERC Discovery Grant program. All opinions, findings, conclusions and recommendations in this paper are those of the authors and do not necessarily reflect the views of the funding agencies.


\end{document}


\section{Appendix - More Training Details}\label{sec:train_detail}

\subsection{Speed Test for Different Models}

\begin{table}[h]
\small
\centering
\begin{tabular}{lcc}  
\toprule
\multirow{2}{*}\textbf{{Model}}           & \textbf{Training Speed}  & \textbf{Inference Speed} \\
                & (QPS)                 & (QPS) \\
\midrule
BERT$_{12}$     & 2.69 (1.00$\times$)  & 7.62 (1.00$\times$) \\
RoBERTa$_{12}$  & 2.54 (0.94$\times$)  & 7.54 (0.99$\times$)  \\
XLNet$_{12}$    & 5.45 (2.03$\times$)  & 3.44 (0.45$\times$)  \\
ALBERT$_{12}$   & 2.10 (0.78$\times$)  & 6.02 (0.79$\times$)  \\
\midrule
BERT$_6$-FT     & 6.34 (2.36$\times$)  & 23.93 (3.14$\times$) \\
BERT$_6$-V-KD   & 4.13 (1.54$\times$)  & 23.28 (3.06$\times$) \\
BERT$_6$-RL-KD  & 2.66 (0.99$\times$)  & 23.67 (3.11$\times$) \\
\midrule
BERT$_3$-FT     & 9.94 (3.70$\times$)  & 45.89 (6.02$\times$) \\
BERT$_3$-V-KD   & 6.79 (2.52$\times$)  & 44.48 (5.84$\times$) \\
BERT$_3$-RL-KD  & 3.28 (1.22$\times$)  & 44.82 (5.88$\times$) \\
\bottomrule
\end{tabular}
\caption{Average training \& inference speed for different approaches on QQP task. Queries Per Second (QPS) means average number of cases processed per second. We use this metric to evaluate the model training \& inference speed.}
\label{tab:runtime}
\end{table}

All the model training experiments are performed on a machine with 4 V100 16G GPUs. Taking QQP task as an example, the training and inference speed test for different models are shown in Table~\ref{tab:runtime}, under the setting with batch size = 32, max sequence length = 128 and epoch = 4. 

In the training process, RL-KD takes longer time than FT and V-KD methods, which is expected due to the joint training of TS and KD models. For model inference (we care more about), there is no big difference between different methods. The motivation of this paper is to distill a smaller student model for faster inference speed. Although our approach may take longer time for model training, the student model still achieves very decent inference speed with much better results compared with baseline distillation approaches. 


\subsection{Model Parameter Size}

\begin{table}[h]
    \centering
    \small
    \begin{tabular}{lcc}
    \toprule
  \textbf{ Model} &	\textbf{KD Parameter} & \textbf{TS Parameter}\\
   \midrule
   BERT$_{12}$	& 110M  & - \\
   Robert$_{12}$ &	125M & - \\
   XLNet$_{12}$	& 110M & - \\
   ALBERT$_{12}$ & 12M & - \\
   BERT${_6}$ & 67.0M & 3.1k \\
   BERT${_3}$ &	45.7M & 3.1k \\
    \bottomrule
    \end{tabular}
    \caption{Number of parameters in each model.}
    \label{tab:stats_parameter}
\end{table}

Compared with the BERT$_{12}$, the student model (BERT${_3}$, BERT${_6}$) can greatly reduce the parameter magnitude by 58.5\% and 39.1\% respectively. In our approach RL-KD, although we further introduce a teacher selector model for teacher selection, we will not add too many parameters in our model (the amount of added parameters can be ignored compared with KD model). In addition, the size of teacher selector parameters does not depend on the parameter size of student model, but on the number of teacher models and number of the task golden labels. The parameter size for TS model is only 3.1k, compared with 45.7M/67.0M for KD model. 
